%% file: main.tex
\documentclass[letterpaper,twocolumn,10pt, dvipsnames]{article}
\usepackage{usenix2019_v3}
\usepackage{soul}
\usepackage{tikz}
\usepackage{amsmath}
\usepackage{algorithm}
\usepackage{algorithmic}
\usepackage{url}
\usepackage{fancyvrb}
\usepackage{listings}
\usepackage{xcolor}
\usepackage{multirow}
\usepackage{booktabs}
\usepackage{caption}
\usepackage{amsmath}
\usepackage{subcaption}

\newcommand{\sa}[1]{{\color{black} #1}}

\usepackage{colortbl}
\definecolor{lightgray}{gray}{0.9}

\definecolor{codegreen}{rgb}{0,0.6,0}
\definecolor{codegray}{rgb}{0.5,0.5,0.5}
\definecolor{codepurple}{rgb}{0.58,0,0.82}
\definecolor{backcolour}{rgb}{0.95,0.95,0.92}

\usepackage{filecontents}

\lstdefinestyle{mystyle}{
    backgroundcolor=\color{backcolour},   
    commentstyle=\color{codegreen},
    keywordstyle=\color{magenta},
    numberstyle=\tiny\color{codegray},
    stringstyle=\color{codepurple},
    basicstyle=\ttfamily\footnotesize,
    breakatwhitespace=false,         
    breaklines=true,                 
    captionpos=b,                    
    keepspaces=true,                 
    numbers=left,                    
    numbersep=5pt,                  
    showspaces=false,                
    showstringspaces=false,
    showtabs=false,                  
    tabsize=1,
    frame=tb, 
    frameround=tttt,
    linewidth=1\linewidth, 
    columns=fullflexible, 
}

\AtBeginDocument{%
  \providecommand\BibTeX{{%
    \normalfont B\kern-0.5em{\scshape i\kern-0.25em b}\kern-0.8em\TeX}}}

\input{header}
\input{terminologies}

\iftrue
\newcommand{\sm}[1]{\todo[inline,color=brown!40]{Subrata: #1}}

\else
\newcommand{\sm}[1]{}
\newcommand{\sa}[1]{}
\newcommand{\sarc}[1]{}

\fi

\lstdefinestyle{mystyle}{
    backgroundcolor=\color{gray!20},
    basicstyle=\footnotesize\ttfamily,
    frame=single,
    framerule=0.5pt, 
    framesep=2pt, 
    rulecolor=\color{gray!50}, 
}

\begin{document}
\date{}






\title{\vspace{-2em}Prompt-Aware Scheduling for Efficient Text-to-Image Inferencing System\vspace{-1em}}

 
\author{
{\rm Shubham Agarwal \thanks {Poster presented at NSDI'24}}\\
Adobe Research \\
\textit{shagarw@adobe.com}
\and
{\rm Saud Iqbal}\\
Adobe Research \\
\textit{saudi@adobe.com}
\and
{\rm Subrata Mitra}\\
Adobe Research \\
\textit{sumitra@adobe.com}
\vspace{-5em}
} 







\maketitle

\input{tex/abstract.tex}

\input{tex/introduction.tex}

\input{tex/overview.tex}
\input{tex/evaluation.tex}
\input{tex/conclusion.tex}

{\footnotesize

}

\end{document}

%% file: header.tex
\usepackage{kantlipsum,setspace,titlesec}
\usepackage{todonotes}

\usepackage{tikz}
\usepackage{amsmath}
\usepackage{amsthm}
\usepackage{amssymb}
\usepackage{xspace}
\usepackage{bm}
\usepackage{xurl}
\usepackage{filecontents}
\usepackage{cleveref}
\usepackage{comment}
\usepackage{graphicx}
\usepackage[font={bf,it,footnotesize},labelsep=colon, belowskip=-1.0pt, aboveskip=0.3pt]{caption}
\usepackage[shortlabels]{enumitem}
\setlist[enumerate]{nosep}
\setlist{nolistsep,leftmargin=4.0mm}
\usepackage{mathtools}
\usepackage[hang,flushmargin]{footmisc} 
\usepackage{xcolor}
\usepackage{algorithm}
\usepackage{algorithmic}
\usepackage{kantlipsum,setspace,titlesec}
\usepackage{subcaption}

\titlespacing{\section}{0em}{1em}{1em}
\titlespacing{\subsection}{0em}{0.5em}{0.4em}
\titlespacing{\subsubsection}{0em}{0.1em}{0.1em}

\titlespacing{\section}{0pt}{2ex}{1ex}
\titlespacing{\subsection}{0pt}{1ex}{0ex}
\titlespacing{\subsubsection}{0pt}{0.5ex}{0ex}

\usepackage{xpatch}

%% file: terminologies.tex
\newcommand{\AC}[0]
{\texttt{ApproxC}\xspace}



%% file: tex/abstract.tex
\begin{abstract}
\vspace{-1em}
Traditional ML models utilize controlled approximations during high loads, employing faster but less accurate models in a process called \textit{accuracy scaling}. However, this method is less effective for generative text-to-image models due to their sensitivity to input prompts and performance degradation caused by large model loading overheads. This work introduces a novel text-to-image inference system that optimally matches prompts across multiple instances of the same model operating at various approximation levels to deliver high-quality images under high loads and fixed budgets.
\end{abstract}

%% file: tex/introduction.tex
\section{Introduction}

Text-to-image generation using Diffusion Models has become very popular and is being offered by various companies~\cite{firefly}. However, serving diffusion models pose challenges as they use 50 to 100 denoising steps which take up to 5 seconds even on A100 GPUs~\cite{podell2023sdxl}. High-throughput inference-serving systems like \cite{ahmad2024proteus,crankshaw2017clipper} employ multiple ML models with different accuracy-latency-cost trade-offs to handle incoming load. However, their scalability for diffusion models is limited due to several drawbacks: (1) Existing systems switch to less accurate (faster) models under high load, assuming input-agnostic accuracy measures. However, for text-to-image models, image quality can vary significantly based on the input prompt. (2) Model switching overhead is significant for large models like Diffusion Models (3) Horizontal scaling to meet varying query demands can be expensive and unreliable.

Alternatively, Agarwal et al. introduced a novel system~\cite{agarwal2023approximate, lu2024recon} to decrease generation latency by employing \textit{approximate caching} (\AC), selectively skipping certain denoising iterations and reusing prior intermediate states based on input prompt closeness with the cache. However, it is a single GPU serving system, and it cannot handle high loads without horizontal scaling. Also, the \AC technique being input prompt dependent, indiscriminate skipping of iterations under high-load can lead to very bad quality of output.

For building a high throughput text-to-image serving system, it should try to consciously align each prompt with the most suitable approximate model for high quality and also eliminate the overhead tied to model loading and unloading. At a high level, the inferencing system employs two design propositions. First, it micromanages prompt-to-model allocation to maintain image quality under varying loads. Second, it utilizes \AC to adjust a single model's latency and accuracy for varying loads, avoiding any switching overheads.



\begin{figure}[t]
\centering
        \includegraphics[width=0.95\linewidth]{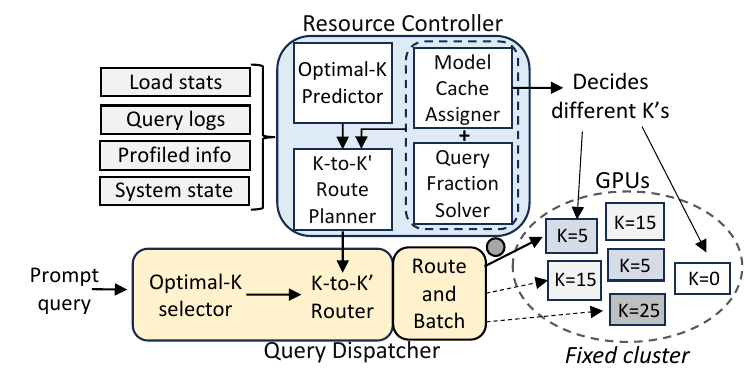}
        \caption{Overview of system}
        \label{fig:run_time}
\end{figure}


\section{Approach}
\noindent\textbf{Overview:} The system operates a fixed-sized GPU inference serving cluster~\cite{ahmad2024proteus}, but with a unique approach. Instead of employing multiple model variants, it runs the same model on all GPUs and employs \AC~\cite{agarwal2023approximate} to balance accuracy and inference latency trade-off.  
First, the system solves an optimization problem based on system load to determine the number of model instances and their respective $K$ values (the number of initial denoising iterations skipped in \AC) and also calculate the fraction of the input load allocated to each instance to ensure high throughput at a macro-level. Then, at a micro-level, it uses a heuristic to assign input prompts to specific model instances with particular \textit{optimal-$K$} values (the least number of inference steps to generate optimal quality image). This ensures that while the macro-level allocation requirements are met, at the micro-level, the system achieves optimal prompt-to-$K$ matching to enhance generation quality. However, it might be unable to assign all incoming prompts to its  \textit{optimal model} $K$. Hence, to accommodate the load, it judiciously redirects a prompt to a model running at a different $K'$ using a redirection logic, which aims to minimize quality degradation.  Furthermore, to enhance throughput, the system uses a tailored \textit{route-and-batch} technique.

%% file: tex/overview.tex
\noindent\textbf{Resource Controller:} 
The \texttt{Controller} runs periodically using query logs, workload, and system state information. Using the optimization problem outlined in \cite{ahmad2024proteus}, the \texttt{Model Cache Assigner} determines the optimal distribution of models across different values of \(K\) for \AC, and the \texttt{Query Fraction Solver} calculates the proportion of prompts to be redirected to model at \(K\) for runtime query flow to effectively manage the load, denoted by \(F(K)\). Additionally, the \texttt{Optimal-K Predictor} forecasts the optimal-K distribution (\(H_K\)) for the incoming prompt queries to maintain quality.
Since \(H_K\) and \(F_K\) may differ, the prompts may be redirected to models running at \(K\) values different from their optimal-\(K\) values. To address this, the \texttt{K-to-K' Route Planner} is designed to find redirection probabilities aimed to sustain throughput while maintaining quality. It aims to minimize the quality degradation \(\mathcal{D}_Q\) (in Eq. 1) to determine which prompts should be redirected to which GPU based on their predicted affinity for an optimal-\(K\) and the available GPUs running at certain \(K'\) values, thus providing a \textit{Route-Plan}.  This \textit{Route-Plan} is used by the \texttt{Query Dispatcher} as \textit{Redirection Logic} to assign incoming prompts to appropriate \texttt{Workers} running at \(K\). For an incoming prompt with an optimal-\(K\), this \textit{Route-Plan} determines the appropriate alternate value of \(K\) (referred to as \(K'\)) to be used under the present load situation. It can shift queries either to a slower/better model running at \(K'\) such that \(K' < K\), or to the \textit{closest} possible faster/worse model running at \(K'\) such that \(K' > K\) (with quality degradation $D$), while minimizing overall quality degradation (\(\mathcal{D}_Q\)).

\begin{minipage}{0.45\textwidth}
{
\centering
\small
\begin{equation}\label{eq:minimize}
\text{Minimize } \mathcal{D}_Q = \sum_{i,j \text{ s.t.}}\sum_{K'_{j} > K_i} P(K'_{j}|K_i) \cdot \mathit{H}_k(K_i) \cdot \mathit{D}(K'_{j}, K_i)
\end{equation}
}
\end{minipage}

\noindent\textbf{Query Dispatcher:} The \texttt{Query Dispatcher} directs prompts to appropriate \AC models (based on the idea outlined in~\cite{agarwal2023approximate}) running at \(K\) on \texttt{GPU workers}. 
To achieve this, the \texttt{Optimal-K Selector} first retrieves the nearest cache and determines the optimal \(K\). Subsequently, the \texttt{K-to-K' Router} selects the final approximate model at \(K'\) based on the \textit{Route-Plan} computed by the \texttt{Controller}.

It utilizes a specialized \textit{load-aware} \textit{route-and-batch} approach, alternating between \textit{uniform} and \textit{greedy} routing based on load. During low loads, it employs \textit{uniform} routing with a batch size of 1, distributing prompts randomly to workers (running at $K$). Conversely, at high loads, \textit{greedy} routing assigns prompts to \texttt{GPU workers} with the longest queues to maximize throughput using optimal batch size. This strategy optimizes latency under low loads and employs the carefully designed routing and batching technique to increase throughput and reduce SLO violations under high loads by selecting the worker likely to be fired soonest at optimal batch size.

\begin{figure}[t]
\centering
\begin{minipage}[b]{0.65\columnwidth}
  \centering
  \includegraphics[width=\linewidth]{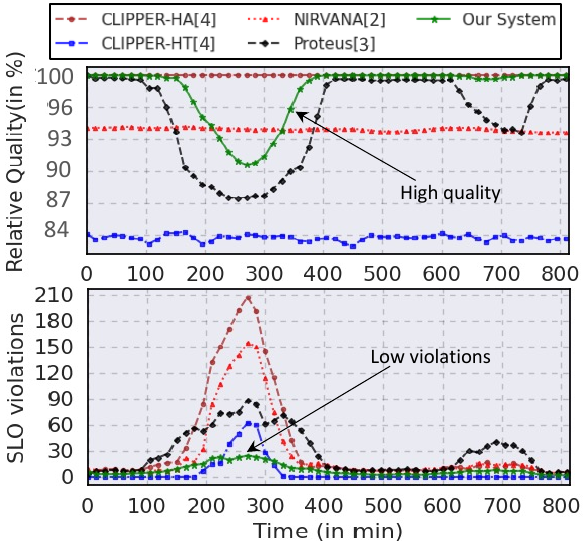}
  \caption{Performance of system on Twitter trace}
  \label{fig:eval}
\end{minipage}%
\hfill
\begin{minipage}[b]{0.35\columnwidth}
  \centering
  \includegraphics[width=0.95\linewidth]{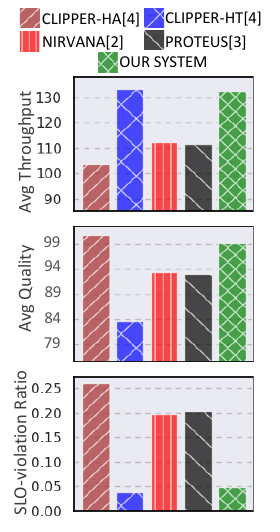}
  \caption{Aggregated}
  \label{fig:eval_second}
\end{minipage}
\end{figure}

%% file: tex/evaluation.tex
\section{Evaluation and Discussion}
\label{sec:eval}

We assessed our system using SD-XL~\cite{podell2023sdxl} models on an 8 NVIDIA A100 GPU setup, combining production and synthetic workloads with prompts from DiffusionDB~\cite{wang2022diffusiondb}. In the Twitter trace workload (Fig. \ref{fig:eval} and Fig. \ref{fig:eval_second}), \textit{Clipper-HA}~\cite{crankshaw2017clipper} achieves near-perfect relative quality but suffers the highest SLO violations (25\%). Conversely, \textit{Clipper-HT}~\cite{crankshaw2017clipper} has lower SLO violations (5\%) and higher throughput (30\%) at the cost of quality (at just 85\%). \textit{NIRVANA}~\cite{agarwal2023approximate} maintains around 94\% average quality but struggles with throughput and SLO violations (20\%) as it can not scale at high workloads. \textit{Proteus}~\cite{ahmad2024proteus} performs well at stable workloads but faces SLO violations (25-30\%) during changes and offers subpar quality (< 90\%) due to prompt-agnostic variant selection and model loading overheads. In contrast, \textit{our system} maintains consistent quality (> 90\%) and achieves the lowest SLO violation ratio (< 5\%) by using prompt-aware variant selection and leveraging \AC variants. Overall, it delivers up to \textbf{10\%} higher quality and up to \textbf{40\%} higher throughput with up to \textbf{10x} lower latency SLO violations compared to baselines.

%% file: tex/conclusion.tex
\textbf{Ongoing works} In this work, we introduced a text-to-image inferencing system that can significantly improve the quality of results, even under high load, by using a novel algorithm to optimally match the prompts across a set of model instances running at different approximation levels. Our current focus includes extending the serving infrastructure to other generative model families and leveraging heterogeneous serving environments with multiple model families and device types.

\section{Conclusion}

We developed and implemented a high-performance inference serving system for text-to-image models, designed to enhance result quality, even during high traffic, through an innovative algorithm that optimally aligns prompts across multiple model instances operating at varying levels of approximation. Additionally, by strategically applying a recent technique known as \textit{approximate caching} and devising an effective batching strategy, our system eliminates the costs associated with model-switching as workload characteristics evolve and minimizes violations of latency SLOs.